\pdfoutput=1

\documentclass[11pt]{article}

\usepackage[preprint]{acl}
\usepackage{amssymb}
\usepackage{colortbl}
\usepackage{times}
\usepackage{latexsym}
\usepackage{tabularx}
\usepackage{booktabs}
\usepackage{amsmath}
\usepackage{float}
\usepackage{multirow}
\usepackage[T1]{fontenc}

\usepackage[utf8]{inputenc}

\usepackage{microtype}

\usepackage{inconsolata}

\usepackage{graphicx}

\title{\textsc{LOOK-M}: Look-Once Optimization in KV Cache \\ for Efficient Multimodal Long-Context Inference}

\author{Zhongwei Wan\textsuperscript{1}$^{\dag}$\thanks{ Work was done at Tencent AI Lab.},  Ziang Wu\textsuperscript{2}\thanks{Equal contribution.}, Che Liu\textsuperscript{3}, Jinfa Huang\textsuperscript{2}, Zhihong Zhu\textsuperscript{2},\\ 
{\bf Peng Jin\textsuperscript{2}}, 
{\bf Longyue Wang\textsuperscript{4}}\thanks{Corresponding authors.}, {\bf Li Yuan\textsuperscript{2}}$^{\ddag}$ \\
\textsuperscript{1}The Ohio State University \quad
\textsuperscript{2}Peking University \\
\textsuperscript{3}Imperial College London \quad
\textsuperscript{4}Tencent AI Lab \\
wan.512@osu.edu, \quad ziangwu7777@gmail.com, \quad che.liu21@imperial.ac.uk \\
\{jinfahuang, jp21, zhihongzhu\}@stu.pku.edu.cn, \quad vinnylywang@tencent.com \\
yuanli-ece@pku.edu.cn\\
Code: \url{https://github.com/SUSTechBruce/LOOK-M}. \\
}

\begin{document}
\maketitle
\begin{abstract}


Long-context Multimodal Large Language Models (MLLMs) demand substantial computational resources for inference as the growth of their multimodal Key-Value (KV) cache, in response to increasing input lengths, challenges memory and time efficiency. 
Unlike single-modality LLMs that manage only textual contexts, the KV cache of long-context MLLMs includes representations from multiple images with temporal and spatial relationships and related textual contexts. The predominance of image tokens means traditional optimizations for LLMs' KV caches are unsuitable for multimodal long-context settings, and no prior works have addressed this challenge.
In this work, we introduce \textbf{\textsc{LOOK-M}}, a pioneering, fine-tuning-free approach that efficiently reduces the multimodal KV cache size while maintaining performance comparable to a full cache. We observe that during prompt prefilling phase, the model prioritizes more textual attention over image features, and based on the multimodal interaction observation, a new proposed text-prior method is explored to compress the KV cache. Furthermore, to mitigate the degradation of image contextual information, we propose several compensatory strategies using KV pairs merging. \textbf{\textsc{LOOK-M}} demonstrates that with a significant reduction in KV Cache memory usage, such as reducing it by \textbf{80\%} in some cases, it not only achieves up to \textbf{1.5x} faster decoding but also maintains or even \textbf{enhances} performance across a variety of long context multimodal tasks.

\end{abstract}

\section{Introduction}

Large language models (LLMs)~\citep{achiam2023gpt, meta2024introducing, jiang2023mistral, wan2023efficient} are progressively evolving into multimodal large language models (MLLMs)~\cite{Yang2023TheDO, Yin2023ASO}, making significant advances in the processing of extensive multimodal contexts such as GPT-4V. 
Despite the impressive capabilities of MLLMs, they still face significant challenges when dealing with long multimodal context inputs, such as temporal multi-image tasks and semantic multi-image tasks~\cite{Song2024MileBenchBM}, or multi-turn multimodal dialogues~\cite{team2023gemini} in real-world applications. Specifically, multimodal KV caches hinder the efficient processing of long multimodal inputs. During inference, the increased lengths of inputs linearly slow down the decoding process due to the attention computations across past multimodal KVs.

\begin{figure}[t]
    \centering
    \includegraphics[width=0.485\textwidth]{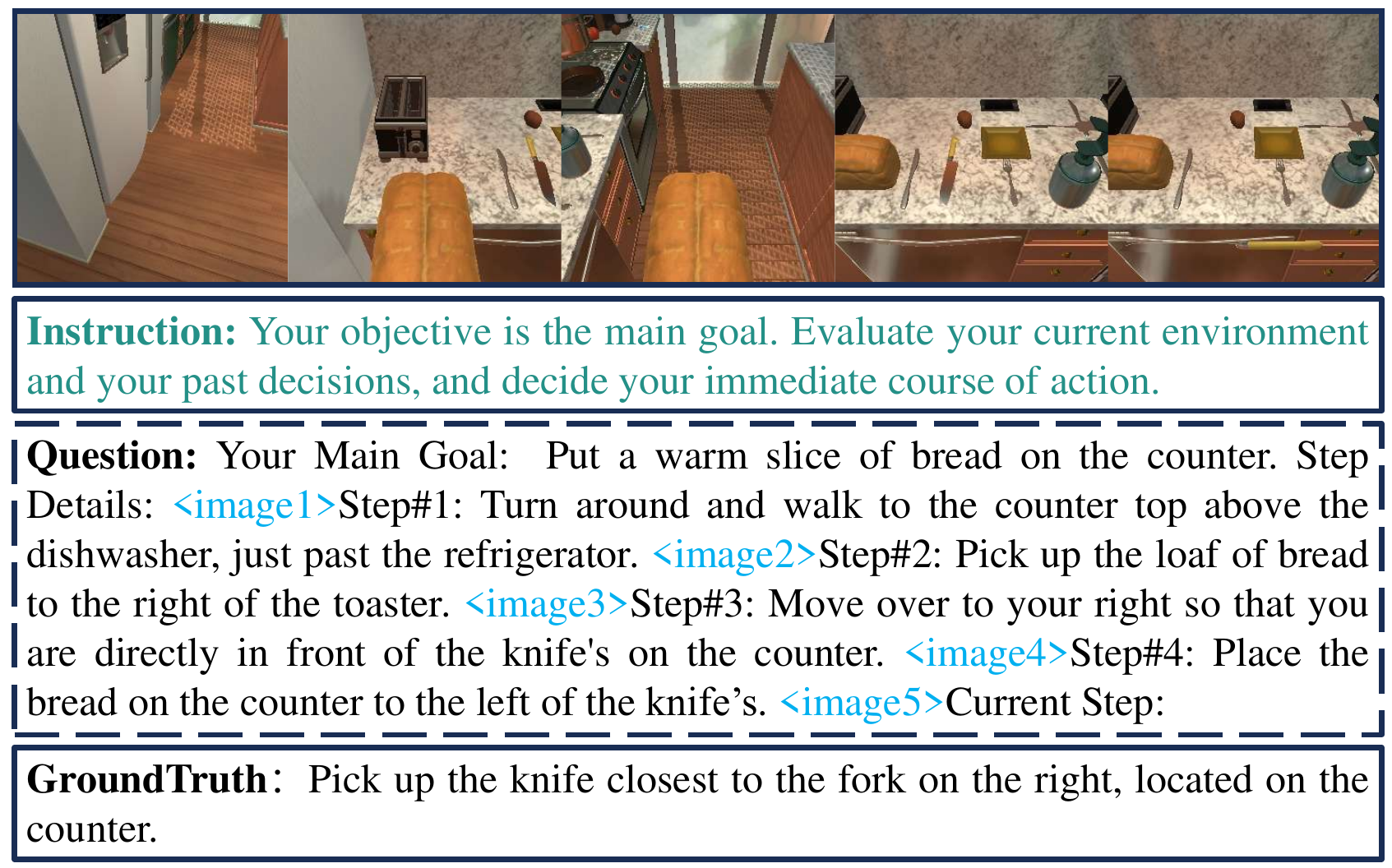}
    \vspace{-0.1in}
    \caption{\small{ A multimodal long-context sample contains multiple images from MileBench~\cite{Song2024MileBenchBM} showing comprehensive spatial relationships.}}
    \vspace{-0.27in}
    \label{fig: sample1}
\end{figure}

\begin{figure}[t]
    \centering
    \includegraphics[width=0.485\textwidth]{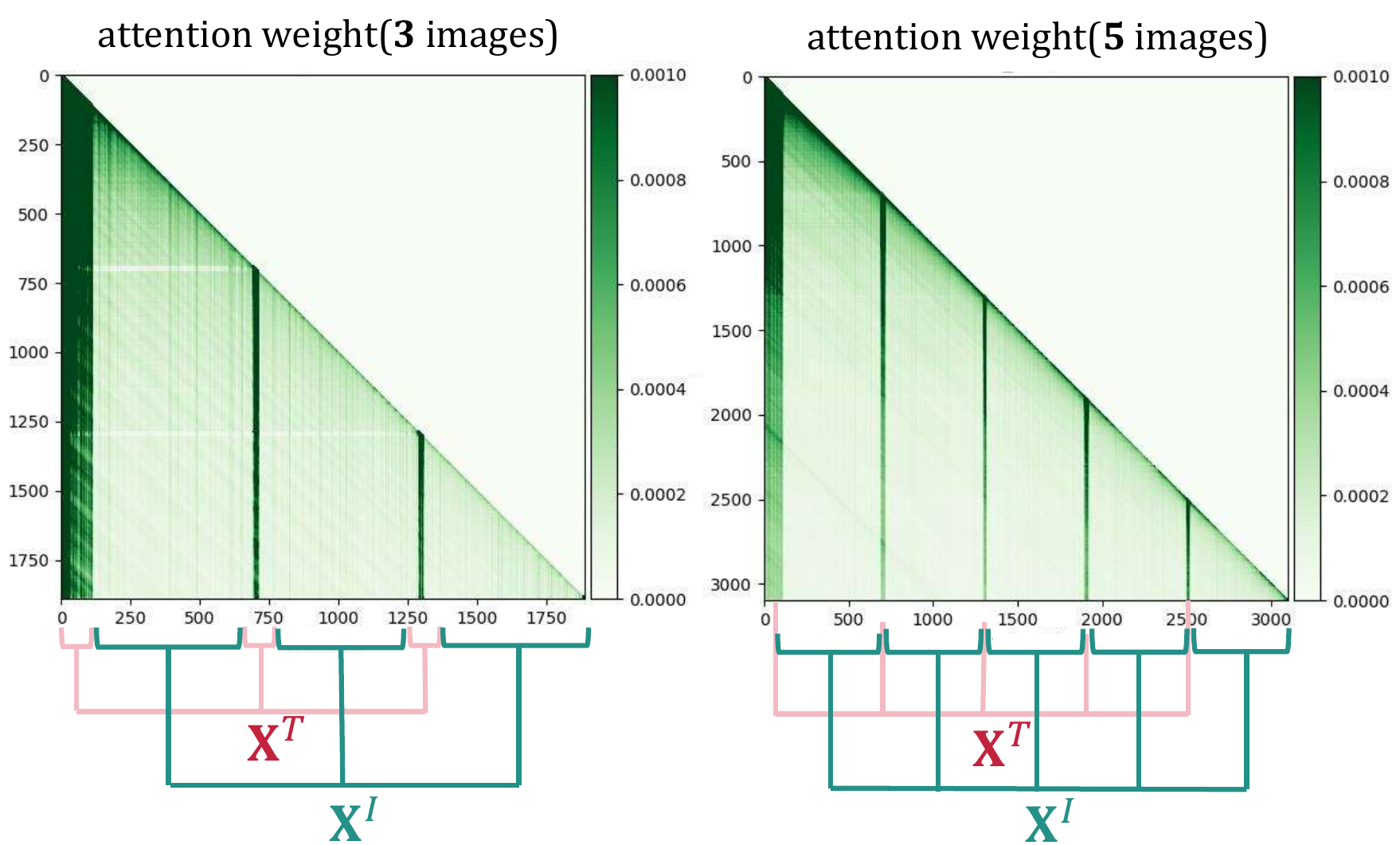}
    \vspace{-0.3in}
    \caption{\small{ 
Visualization of attention in multimodal prompt encoding phase, where $\mathbf{X}^{T}$ represents a text sentence and $\mathbf{X}^{I}$  denotes a subsequent image, showcasing the interleaved input of text and images in multimodal long-context scenarios.}}
    \vspace{-0.2in}
    \label{fig: sample2}
\end{figure}

Furthermore, as depicted in Figure~\ref{fig: sample1}, in contrast to text-only LLMs' KV cache eviction methods~\cite{Zhang2023H2OHO, wan2023efficient}, long multimodal inputs typically include multiple interrelated images, along with definitions or background descriptions relevant to the task. Directly applying traditional text-centric KV cache eviction strategies~\cite{Zhang2023H2OHO, Ge2023ModelTY, Ren2024OnTE, Li2024SnapKVLK} to MLLMs overlooks the potential interactions between multimodal representations~\cite{team2023gemini}.  Specifically, Figure~\ref{fig: sample2} shows the attention visualization for multimodal long-context,  the model exhibits greater attention to the textual components during the multimodal prompt encoding process. This observation demonstrates that the model tends to understand global visual content through textual knowledge, highlighting the necessity of preserving textual features and selectively pruning redundant image tokens in the multimodal KV cache to maintain the integrity of the multimodal context.

In this paper, we introduce \textbf{\textsc{LOOK-M}}, a pioneering and efficient framework that marks the first effort to compress KV caches specifically for multimodal long-context scenarios. The term \textbf{Look-Once} in our method implies that pruning occurs only once during multimodal long prompt encoding, and the model effectively sees the full image just once. LOOK-M utilizes a text-prior technique that prioritizes the retention of textual KV pairs during the prompt encoding phase, given the insight from  Figure~\ref{fig: sample2}. For visual representation, inspired by attention-based eviction strategies~\cite{zhang2024h2o}, our method prunes redundant visual KV pairs that show sparse patterns in attention visualizations, utilizing the metric of attention scores. Furthermore, to preserve global contextual information in the compressed cache, we develop several merging strategies to merge the evicted KV tokens into conserved ones, addressing potential hallucinations and contextual inconsistencies~\cite{Yang2024NoTL} during the decoding process.

Remarkably, LOOK-M does not require any fine-tuning and can be applied in a plug-and-play manner with a look-once KV cache compression strategy.  We evaluate our LOOK-M with several strategies over four recent MLLM backbones LLaVA-v1.5-7B/13B~\cite{Liu2023VisualIT}, MobileVLM-v2~\cite{Chu2024MobileVLMVF} and InternVL-v1.5~\cite{Chen2023InternVLSU} across several multimodal long-context tasks from MileBench~\cite{Song2024MileBenchBM}: temporal multi-image tasks, semantic multi-image tasks, needle in a haystack task, and image retrieval tasks, respectively. Compared to baselines, LOOK-M achieves minimal performance drop with a fixed KV cache budget and improves the model inference decoding latency by \textbf{1.3x} to \textbf{1.5x} and reduces KV Cache memory footprint by \textbf{80\%} to \textbf{95\%} while still maintaining performance on long context multimodal tasks, and even showing improved performance across various tasks. Our analysis validates that combining text-prior and proposed merging strategies contributes to
the multimodal KV cache compression effectiveness of LOOK-M.

\begin{figure*}[ht]
    \centering
    \includegraphics[width=1\textwidth]{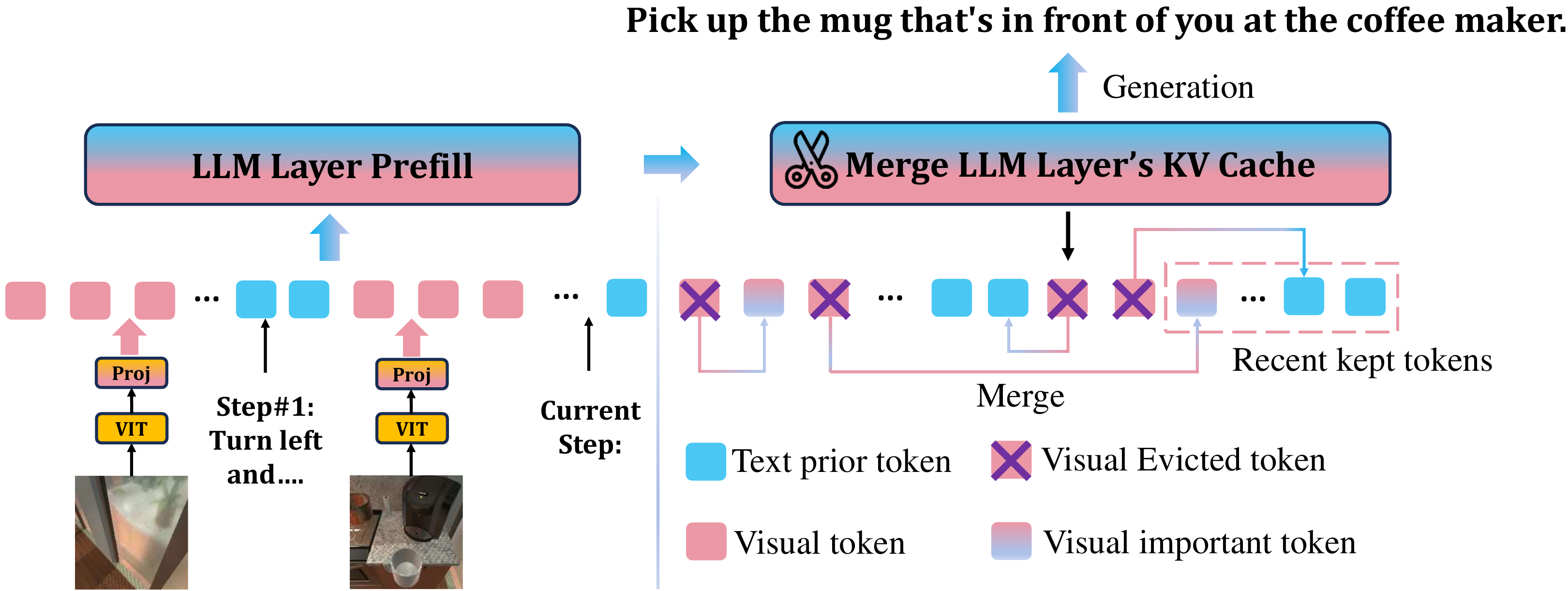}
    \vspace{-0.1in}
    \caption{\small{ 
Pipeline of LOOK-M's KV cache optimization strategy. `Prefill' denotes prompt encoding.}}
    \vspace{-0.1in}
    \label{fig: samplen}
\end{figure*}

\section{Related work}

\noindent \textbf{Vision Token Compression For MLLMs.} 
Classical works in this category, including MobileVLM~\cite{chu2024mobilevlm}, LLaVA-Prumerge~\cite{Shang2024LLaVAPruMergeAT}, MADTP~\cite{Cao2024MADTPMA}, and FastV~\cite{Chen2024AnII}, focus on reducing the number of image tokens, which constitute the majority of total tokens. These methods enhance inference speed by eliminating redundant image tokens. Specifically, MobileVLM~\cite{chu2024mobilevlm} employs a lightweight projector architecture featuring an average pooling layer to significantly compress the number of visual tokens. LLaVA-Prumerge~\cite{Shang2024LLaVAPruMergeAT} and MADTP~\cite{Cao2024MADTPMA} introduce adaptive approaches to visual token reduction, effectively decreasing their count while maintaining model performance. FastV~\cite{Chen2024AnII} introduces a versatile plug-and-play method that optimizes computational efficiency through adaptive attention patterns in early layers and visual token pruning in later stages, achieving up to a 45\% reduction in computational costs while preserving performance.
Unlike these methods, which focus solely on optimizing VIT output tokens and require fine-tuning, LOOK-M specifically targets multimodal token compression within the KV cache without necessitating additional fine-tuning.

\text{ }

\noindent \textbf{KV Cache Compression For LLMs. } 
KV cache compression primarily encompasses three strategies: Eviction, Quantization, and Trainable Compression. In eviction, techniques like Mistral-7B~\cite{jiang2023mistral} and StreamingLLM~\cite{Xiao2023EfficientSL} only preserve key tokens for efficient sequence generation, while approaches like $\text{H}_{2}\text{O}$\cite{zhang2024h2o} and SnapKV~\cite{Li2024SnapKVLK} focus on maintaining a small, influential set of tokens to enhance performance, though risk losing context with evicted KVs. Quantization strategies such as KIVI~\cite{liu2024kivi} and Gear~\cite{kang2024gear} reduce cache memory through advanced quantization techniques, balancing memory efficiency with precision. In trainable Compression, methods like LESS~\cite{dong2024get} and DMC~\cite{nawrot2024dynamic} adapt LLMs to compress KV caches by training on selected datasets, although they face challenges in generalization. However, our LOOK-M utilizes a plug-and-play approach that does not require additional training, ensuring wider applicability without the necessity for tuning specific to multimodal datasets. Therefore, different from these text-centric KV cache compression methods, our \textsc{LOOK-M} specifically targets long multimodal text scenarios and seeks to leverage attention map interactions between text and images to guide KV cache pruning.

\vspace{-0.2mm}
\noindent \textbf{Token Merging. }  Unlike token pruning~\cite{Tang2023DynamicTP, Kong2021SPViTEF, Song2022CPViTCV, Yun2024FocusOT} in encoder-based backbones like ViT~\cite{DBLP:conf/iclr/DosovitskiyB0WZ21} or Bert~\cite{Devlin2019BERTPO}, which discards less significant tokens, token merging~\cite{bolya2022token} consolidates tokens into fewer, more meaningful units, preserving information integrity. Consequently, token merging has become preferred over token pruning to reduce token count. Existing methods like TPS~\cite{wei2023joint}, MG-ViT~\cite{zhang2024mg}, and PuMer~\cite{Cao2023PuMerPA} have explored token merging and pruning techniques, primarily in computer vision tasks. In contrast, \textsc{LOOK-M} is a pioneering effort to adapt token merging within the multimodal KV cache in long-context scenarios, enhancing efficiency for auto-regressive tasks in MLLMs.

\section{Methodology}
In Section~\ref{sec: Background}, we first review the basic implementation of generative inference utilizing a multimodal KV cache. Subsequently, as shown in Figure~\ref{fig: samplen}, we detail the principal components of the LOOK-M model, which includes text-prior KV pairs eviction strategy to facilitate precise pruning, discussed in Section~\ref{Sec: text-prior-eviction}, and various strategies for merging KV pairs, such as averaged, pivotal, and weighted merging in Section~\ref{Sec: KV merging}.

\subsection{Preliminary: Generative Inference with Multimodal KV Cache}
\label{sec: Background}
 
A typical generative inference process for MLLMs involves encoding multimodal prompts and generating tokens.

\noindent \textbf{Multimodal Prompt Encoding}. 
During the prompt encoding phase, a sequence of prompts is used to construct a KV cache for each transformer layer in MLLMs. Consider the input prompt tensor $\mathbf{X} \in \mathbb{R}^{L_{\text{prompt}} \times D}$, represented as $\mathbf{X} = \{\mathbf{X}^{T}_{1}, \mathbf{X}^{I}_{1}, \ldots, \mathbf{X}^{T}_{N}, \mathbf{X}^{I}_{M}\}$, where $\mathbf{X}^{T}$ and $\mathbf{X}^{I}$ denote textual and visual embeddings, and $M$ and $N$ represent the number of image and text representations, respectively. Here, $L_{\text{prompt}}$ indicates the prompt length and $D$ is the model's hidden dimension. In most long multimodal context settings, $\mathbf{X}^{T}$ and $\mathbf{X}^{I}$ are interleaved as inputs.
 For simplicity, the indices for heads and layers have been omitted. The key and value tensors are derived as follows:
\begin{equation}
    \mathbf{K}=\mathbf{X} \mathbf{W}_K, \mathbf{V}=\mathbf{X} \mathbf{W}_V,
\end{equation}
With $\mathbf{W}_K, \mathbf{W}_V \in \mathbb{R}^{D \times D}$ representing the weights for the key and value layers, respectively, $\mathbf{K}$ and $\mathbf{V}$ are computed and subsequently stored in the KV cache to aid in token generation.

\noindent \textbf{Token Generation}. During the Token Generation phase, the KV cache is employed and updated to sequentially generate tokens. At each time step $t$, keys and values are computed only for the new token $\mathbf{x}_{i}$, while those for $\mathbf{x}_{<i}$ are retrieved from the cache. Concatenation is denoted as $[\cdot]$. Following this, the cache is updated, and the output for the newly generated token is given as:
\begin{equation}
    \mathbf{K} = [\mathbf{K}, \mathbf{x}_{t} \mathbf{W}_K], \mathbf{V} = [\mathbf{V}, \mathbf{x}_{t} \mathbf{W}_V],
\end{equation}
\begin{equation}
    \mathbf{x}_{t, out} = \operatorname{Softmax}\left(\mathbf{q}_{t} \mathbf{K}^{\top} / \sqrt{D}\right) \mathbf{V}, \mathbf{q}_{t} = \mathbf{x}_{t} \mathbf{W}_Q,
\end{equation}
where $\mathbf{W}_Q \in \mathbb{R}^{D \times D}$ is the weight matrix of the query layer, the linear growth of the multimodal KV cache with each new token notably heightens memory consumption and latency, especially with longer prompts or during token generation, highlighting the need for cache compression.

\begin{figure*}[t]
    \centering
    \includegraphics[width=0.99\textwidth]{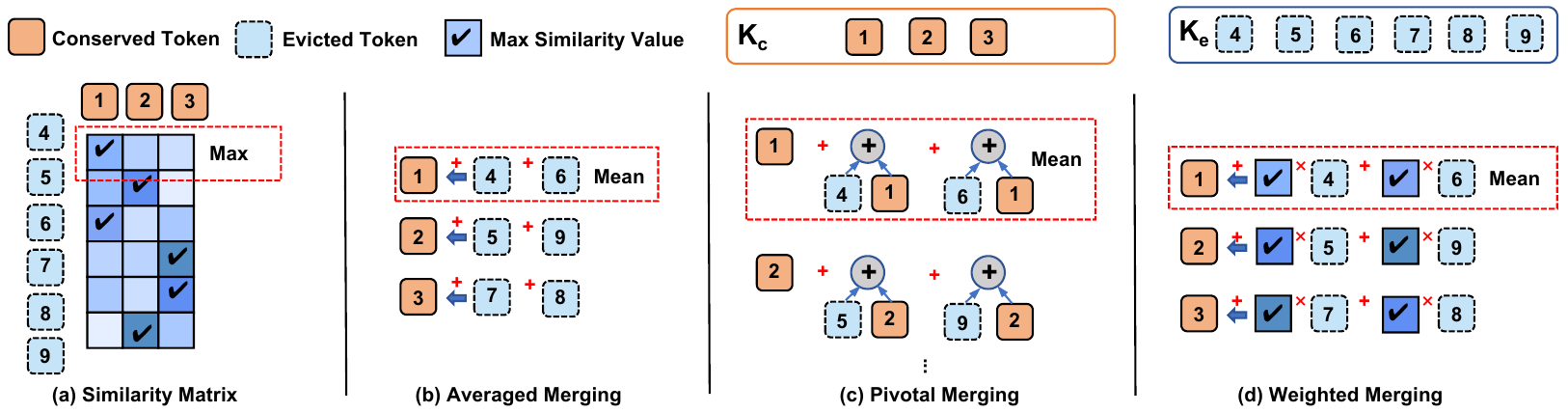}
    \vspace{-0.1in}
    \caption{\small{A simple similarity matrix example and Four merging strategies of LOOK-M: Averaged Merging, Pivotal Merging, and Weighted Merging.}}
    \vspace{-0.1in}
    \label{fig: four_strategies}
\end{figure*}

\subsection{Text-Prior KV Pairs Eviction}
\label{Sec: text-prior-eviction}

The key idea of KV pair eviction during the prompt prefilling phase is to dynamically update the KV cache using cumulative attention scores. This process strategically excludes the least essential KV pairs to maintain a compact cache size, thereby ensuring that only the most valuable tokens are preserved for efficient inference.
However, contrary to the traditional accumulation-based approach ~\cite{zhang2024h2o} that will indiscriminately treat all tokens, our method prioritizes the retention of text-based KV pairs and performs eviction of image-based KV pairs, guided by the patterns observed in the attention visualizations shown in Figure~\ref{fig: sample2}, and then integrating them within a recent window with size $M$. Let $T$ denotes the indices of textual tokens, $\text{T}_{p}$ denotes text-prior value, the attention score $\mathbf{A}_{s}$ is formulated as follows:
\begin{equation}
\mathbf{A}_{s} = \sum_{i=0}^{L_{\text{prompt}}}\mathbf{A}_{p}[i,:], \text{ }\mathbf{A}_{p} = \operatorname{Attn}\left(\mathbf{Q}_{p} \mathbf{K}_{p}^{\top} \right), 
\end{equation}
\begin{equation}
    \mathbf{A}_{s}[T]=\mathbf{A}_{s}[T] + \text{T}_{p}, \text{ T}_{p}=\text{Max}(\mathbf{A}_{s}), 
\end{equation}
where $\mathbf{A}_{p}$ denotes the attention weight of prompt encoding, $\mathbf{Q}_{p}, \mathbf{K}_{p} \in \mathbb{R}^{L_{\text{prompt}} \times D}$. We set $\text{T}_{p}$ as the maximum value of $\mathbf{A}_{s}$ to prioritize text tokens for preservation. 
After calculating the current cumulative attention scores, we preserve the most recent window of size 
$M$. Subsequently, from the remaining KV cache, the top 
$N$ important tokens with the highest scores are selected to finalize the eviction. The process is defined as follows:

\begin{equation}
    \mathbf{K}_{c} = [ \mathbf{K}[I, :], \mathbf{K}[-M:, :]],
\end{equation}
\begin{equation}
    \mathbf{V}_{c} = [\mathbf{V}[I, :], \mathbf{V}[-M:, :]],
\end{equation}
\begin{equation}
    \text{and }  I = \text{Top}_{N}\left(\mathbf{A}_{s}[:-M], N\right),
\end{equation}
where $\text{Top}_{N}\left(\cdot, N\right)$ selects the indices of  top $N$ important tokens in AttnScore, $I$ denotes the union of textual token indices $T$ and the Top $N$ tokens. $\left(\mathbf{K}_{c}, \mathbf{V}_{c}\right)$ is the conserved KV cache after eviction. Therefore, the compressed multimodal KV cache size is $S = N + M$.

\subsection{KV Pairs Merging Strategies}
\label{Sec: KV merging}
To mitigate the loss of context information following the eviction of multimodal KV pairs, we explore various merging strategies during the prompt encoding phase. 
Given the eviction set $\mathbf{K}_{e} = \mathbf{K} - \mathbf{K}_{c}$, 
we deploy a many-to-one nearest-neighbor matching algorithm~\cite{dang2021nearest} to derive the similarity matrix $\mathbf{S}$ between $\mathbf{K}_{e}$ and $\mathbf{K}_{c}$. Considering the alignment properties of KV-pairs in MLLMs, we only compute the similarity matrix on the key's tokens and share the similarity matrix and weighted merging weights with the value's tokens. 
More specifically, $I^{e}$ and $I^{c}$ represent the indices, and $L^{e}$ and $L^{c}$ signify the token lengths in $\mathbf{K}_{e}$ and $\mathbf{K}_{c}$, respectively. Within the matrix $\mathbf{S}$, each element $\mathbf{s}_{i, j}$ captures the interaction required for matching tokens, where $i \in I^{e}$ and $j \in I^{c}$. The process starts by identifying the nearest token $\mathbf{k}^{\text{closest}}$ within $\mathbf{K}_{c}$ for each token $\mathbf{k}_{i}$ from the evicted set. The respective formulas are as follows:
\begin{equation}
    \mathbf{k}_{\mathbf{K}_{c} \rightarrow \mathbf{K}_{e}}^{\text{closest}} = \underset{j \in I^{c}}{\text{Argmax}} \left(\mathbf{s}_{i, j}\right),\text{ } \mathbf{s}_{i, j}=\frac{\mathbf{k}_i^{\top} \mathbf{k}_j}{\left\|\mathbf{k}_i\right\|\left\|\mathbf{k}_j\right\|},
\end{equation}
We utilize cosine similarity where $|\cdot|$ denotes the norm, and matrix $\mathbf{S} \in \mathbb{R}^{L^{e} \times L^{c}}$.. Subsequently, we introduce three novel merging strategies for integrating evicted and conserved KV tokens, namely averaged merging, pivotal merging, and weighted merging.

\textbf{ }

\noindent \textbf{Averaged Merging}
We begin by exploring a straightforward averaged merging strategy. After computing the similarity matrix $\mathbf{S}$ and obtaining the maximum value from each row to identify the $\mathbf{k}_{\mathbf{K}_{c} \rightarrow \mathbf{K}_{e}}^{\text{closest}}$, we observe that each $\mathbf{k}_{c}$ may have a corresponding maximum similarity set $\mathbf{k}_{sim}$ from  $\mathbf{K}_e$, since the relationship between the evicted tokens $\mathbf{K}_e$ and the conserved tokens $\mathbf{K}_c$ is one-to-many. As demonstrated in Figure~\ref{fig: four_strategies} (b), given the results from the similarity matrix, the maximum similarity set for token 1 includes tokens 4 and 8. We employ the most direct method of averaging for the merging:
\begin{equation}
    \mathbf{k}_{c} = \frac{1}{L_{\text{sim}} + 1} (\mathbf{k}_{c} + \sum_{i=0}^{L_{\text{sim}}} \mathbf{k}_{sim}[i]), \text{ }\mathbf{k}_{sim} \in \mathbf{K}_e,
\end{equation}
where $L_{\text{sim}}$ denotes the number of $\mathbf{K}_e$ tokens.

\textbf{ }

\noindent \textbf{Pivotal Merging} Unlike averaged merging, the pivotal merging approach emphasizes the weight proportion for the conserved tokens $\mathbf{K}_{c}$ during the merging process. As illustrated in Figure~\ref{fig: four_strategies} (c), we initially perform an average fusion between each $\mathbf{k}_{e}$ and its corresponding $\mathbf{k}_{\mathbf{K}_{c} \rightarrow \mathbf{K}_{e}}^{\text{closest}}$. The merged tokens are designated as '\textit{pivotal tokens}'. Subsequently, we average merge each $\mathbf{k}_{c}$ with its corresponding pivotal token, as formulated below:
\begin{equation}
    \mathbf{k}_{c} = \frac{1}{L_{\text{sim}} + 1} \{\mathbf{k}_{c} + \frac{1}{2}\sum_{i=0}^{L_{\text{sim}}} (\mathbf{k}_{sim}[i] + \mathbf{k}^{\text{closest}})\}, 
\end{equation}
\textbf{ }

\noindent \textbf{Weighted Merging} Contrast to the static weight allocation strategies used in averaged and pivotal merging, we propose a similarity-based weighted merging method that dynamically allocates weights based on the information in the similarity matrix. Specifically, for each $\mathbf{k}_{c}$ and its corresponding maximum similarity set $\mathbf{k}_{sim}$, weights for the elements in $\mathbf{k}_{sim}$ are dynamically assigned according to the entries in the similarity matrix $\mathbf{S}$, as illustrated in Figure~\ref{fig: four_strategies} (d). Consequently, the formula for weighted merging is as follows:
\begin{equation}
    \mathbf{k}_{c} = \frac{1}{L_{\text{sim}} + 1} \{\mathbf{k}_{c} + \sum_{i=0}^{L_{\text{sim}}} (\mathbf{k}_{sim}[i] \cdot \mathbf{S}[x][y] )\}, 
\end{equation}
where $x$, $y$  represent specific coordinates of each element in $\mathbf{k}_{sim}$ relative to corresponding $\mathbf{k}_{sim}$.

\begin{table*}[t]
\centering
\caption{Performance metrics of various KV Cache Strategy on LLaVA-v1.5-7B/13B on MileBench's tasks with recent ratio $\alpha_{1} = 0.1$ and important ratio $\alpha_{2} = 0.1$. A-Merge, W-Merge, P-Merge denote averaged merging, weighted merging and pivotal merging, respectively. TR represents text-prior KV pairs eviction. }
\label{tab: main_result}
\small 
\begin{tabularx}{\textwidth}{l|*{11}{>{\centering\arraybackslash}X}}
\toprule
Method & T-1 & T-2 & T-3 & T-4 & S-1 & S-2 & S-3 & S-4 & S-5 & NH & IR \\ 
\midrule
\multicolumn{12}{c}{LLaVA-v1.5-7B} \\
\midrule
\rowcolor{gray!16}
\textbf{Full Cache} & 40.0 & 46.0 & 32.2 & 37.8 & 56.9 & 33.3 & 12.6 & 23.4 & 60.5 & 4.7 & 4.3 \\
\midrule
$\textbf{H}_{\textbf{2}}\textbf{O}$~\cite{Zhang2023H2OHO} & 40.2 & 46.0 & 31.8 & 38.5 & 55.0 & 33.8 & 12.6 & 22.8 & 60.0 & 1.4 & 3.7 \\ 
\textbf{SnapKV}~\cite{Li2024SnapKVLK}& 40.0 & 46.0 & 31.5 & \textbf{40.6} & 54.6 & 33.5 & 13.0 & 21.9 & 60.0 & 1.4 & 3.7 \\ 
\textbf{RoCo}~\cite{Ren2024OnTE} & 40.2 & 46.0 & 31.8 & 38.5 & 55.0 & 33.8 & 12.6 & 22.8 & 60.0 & 1.4 & 3.7 \\
\midrule
\textbf{LOOK-M} (\textit{A-Merge}) & \textbf{40.3} & \textbf{46.1} & 32.2 & 39.1 & 54.9 & \textbf{34.0} & 12.9 & 21.4 & \textbf{60.5} & 1.6 & 3.7 \\ 
\textbf{LOOK-M} (\textit{W-Merge}) & \textbf{40.3} & \textbf{46.1} & 31.8 & 39.1 & 55.0 & \textbf{34.0} & \textbf{13.2} & 22.4 & \textbf{60.5} & 1.4 & 3.7 \\ 
\textbf{LOOK-M} (\textit{P-Merge}) & 40.2 & \textbf{46.1} & \textbf{32.5} & 39.8 & 55.1 & 33.8 & 12.9 & 22.5 & \textbf{60.5} & 1.7 & 3.5 \\
\textbf{LOOK-M} (\textit{TP} + \textit{A-Merge}) & 40.2 & \textbf{46.1} & 31.8 & 39.2 & 56.1 & 33.7 & 12.9 & 22.6 & 60.0 & 4.9 & 3.7 \\ 
\textbf{LOOK-M} (\textit{TP} + \textit{W-Merge}) & 40.2 & \textbf{46.1} & 32.0 & 39.0 & 56.5 & 33.8 & 12.9 & 23.1 & 60.0 & 5.1 & 3.5 \\ 

\rowcolor{green!16}
\textbf{LOOK-M} (\textit{TP} + \textit{P-Merge}) & \textbf{40.3} & \textbf{46.1} & \textbf{32.5} & 39.9 & \textbf{57.0} & \textbf{34.0} & 12.8 & \textbf{23.9} & \textbf{60.5} & \textbf{5.3} & \textbf{3.8} \\
\midrule
\multicolumn{12}{c}{LLaVA-v1.5-13B} \\
\midrule
\rowcolor{gray!16}
\textbf{Full Cache} & 39.8 & 46.2 & 30.8 & 48.1 & 64.8 & 48.5 & 13.6 & 28.4 & 60.0 & 12.0 & 1.0 \\ 
\midrule
$\textbf{H}_{\textbf{2}}\textbf{O}$~\cite{Zhang2023H2OHO} & 39.5 & 45.9 & 30.4 & 47.9 & 64.1 & \textbf{48.7} & 13.9 & 25.1 & 59.7 & 3.6 & 0.0 \\ 
\textbf{SnapKV}~\cite{Li2024SnapKVLK} & 39.6 & 46.0 & 30.6 & 47.8 & 64.2 & 48.2 & 13.4 & 22.9 & 59.8 & 4.2 & 1.0 \\ 
\textbf{RoCo}~\cite{Ren2024OnTE} & 39.7 & 45.9 & 30.5 & 48.0 & 64.3 & 48.3 & 13.8 & 24.9 & 59.7 & 3.5 & 0.0 \\ 
\midrule
\textbf{LOOK-M} (\textit{A-Merge})  & 39.7 & 46.1 & 30.7 & 48.0 & 64.6 & 48.0 & 13.3 & 22.1 & 59.8 & 4.6 & 1.0 \\ 
\textbf{LOOK-M} (\textit{W-Merge}) & 39.6 & 46.1 & 30.6 & 47.9 & 64.5 & 48.4 & 13.4 & 23.4 & 59.9 & 4.7 & 1.0 \\

\textbf{LOOK-M} (\textit{P-Merge}) & 39.7 & 46.0 & 30.6 & 48.0 & 64.6 & 48.1 & 13.3 & 25.7 & 59.8 & 5.1 & 1.0 \\ 
\textbf{LOOK-M} (\textit{TP} + \textit{A-Merge}) & 39.7 & \textbf{46.2} & 30.7 & 48.0 & \textbf{65.4} & 48.3 & 13.7 & 26.6 & \textbf{60.0} & 11.2 & 1.0 \\ 
\textbf{LOOK-M} (\textit{TP} + \textit{W-Merge})  & \textbf{39.8} & 46.1 & 30.7 & \textbf{48.1} & 64.8 & 48.2 & 13.9 & \textbf{26.9} & \textbf{60.0} & 11.4 & 1.0 \\ 

\rowcolor{yellow!16} \textbf{LOOK-M} (\textit{TP} + \textit{P-Merge}) & \textbf{39.8} & \textbf{46.2} & \textbf{30.8} & \textbf{48.1} & 65.2 & 48.5 & \textbf{14.1} & 26.6 & \textbf{60.0} & \textbf{11.7} & \textbf{1.0} \\

\bottomrule
\end{tabularx}
\end{table*}

\section{Experiments Setting}

\subsection{Datasets and Metrics}
 MileBench is recognized as the first comprehensive benchmark developed to evaluate Multimodal Long-Length Models (MLLMs) across dimensions of multi-image and extended context, designed to cover a broad spectrum of general scenarios. In this section, we scrutinize the effectiveness of our diverse KV Cache compression strategies across all subtasks of MileBench. The benchmark organizes these into four primary task classifications, denoted as \textbf{T}, \textbf{S}, \textbf{N}, and \textbf{I}, each encompassing a series of specialized sub-tasks:

\noindent \textbf{T}: Temporal Multi-image Tasks, which include four distinct tasks from T-1 to T-4.

\noindent \textbf{S}: Semantic Multi-image Tasks, comprising five sub-tasks, spanning from S-1 to S-5.

\noindent \textbf{N}: Needle in a Haystack Tasks, featuring two specific sub-tasks, N-1 and N-2.

\noindent \textbf{I}: Image Retrieval Tasks, which consists of a single, focused sub-task.

The sub-tasks within MileBench are further divided across various datasets, and we employ evaluation metrics such as Accuracy and ROUGE-L to assess performance. The scores for each sub-task are calculated from the average values of these metrics across the datasets included in that sub-task. For specific details regarding the datasets and their associated metrics, please refer to the Appendix \ref{sec:appendix}, Table \ref{tab:taxonomy_details}.

\subsection{Baselines} To compare the benefits of \textbf{LOOK-M}, we employ the latest KV cache eviction methods as baselines: $\textbf{H}_{\textbf{2}}\textbf{O}$~\cite{zhang2024h2o}, which relies on cumulative attention scores; \textbf{SnapKV}~\cite{Li2024SnapKVLK}, using a pooling strategy; and \textbf{RoCo}~\cite{DBLP:journals/corr/abs-2402-06262}, based on mean attention scores. Notably, these methods are exclusively text-based KV cache compression methods. We utilize their default configurations and adapt them for fair comparison in multimodal long-context scenarios.

\subsection{Implementation Details}
We conducted experiments on  NVIDIA A100 (80GB) and RTX 3090 (24GB) GPUs, employing nine variants of our method to compress the KV Cache of LLaVA-v1.5-7B/13B on ten tasks from MileBench. For all methods, the number of recent tokens size $M$ is $\alpha_{1} \times input\_length$. In addition to the recent tokens, we also retain a number of important token sizes $N$ equal to $\alpha_{2} \times input\_length$, ensuring that at the start of the decoding phase, the memory overhead is ($\alpha_{1} + \alpha_{2}$) proportion that of the original decoding phase, where $\alpha_{1}$ and $\alpha_{2}$ are recent and important ratios. Additionally, our testing process aligns with MileBench's, using the default batch size settings for each dataset.

\section{Experiment Results}
In this section, we present experimental results demonstrating the effectiveness of our LOOK-M strategy for KV cache optimization on the LLaVA-v1.5-7B and 13B~\cite{Liu2023VisualIT}, InternVL-v1.5-7B~\cite{Chen2023InternVLSU}, and MobileVLM\_V2-3B~\cite{chu2024mobilevlm} models. These models were tested across various subtasks of the MileBench dataset~\cite{Song2024MileBenchBM}, highlighting the advantages of our approach in multimodal long-context scenarios. We also examine the impact of KV cache compression on different model architectures, establishing its efficacy across diverse structures. Additionally, we explore how varying KV cache budgets and compression ratios ($\alpha_{1}$ and $\alpha_{2}$) affect model performance. Finally, we assess the computational efficiency of our method by measuring the time and computational load during the decoding phase of compressed models.

\subsection{Main Results on MileBench}
We evaluate the LOOK-M model on the LLaVA-v1.5 7B and 13B using MileBench, as shown in Table~\ref{tab: main_result}. To ensure a fair comparison, we set the recent token ratio $\alpha_{1}$ and the important token ratio $\alpha_{2}$ both at 10\%. The results demonstrate that LOOK-M not only manages multimodal KV cache compression effectively with minimal accuracy impact but also surpasses Full Cache when integrating text-prior and merging strategies, significantly enhancing reasoning accuracy by pruning irrelevant tokens from visual representations. Notably, \textit{TP} + \textit{P-Merge} outperforms text-based KV cache eviction baselines such as $\text{H}_{\text{2}}\text{O}$, SnapKV, and RoCo, indicating that considering attention disparities between text and vision leads to better retention of key information. Moreover, this approach achieves superior outcomes compared to other merging strategies, highlighting the benefits of allocating more weight to conserved tokens in preserving critical information under multimodal KV cache compression. 

Since the \textit{TP} + \textit{P-Merge} strategy achieves the best performance, we use it as the default merging strategy in the following experiments.

\begin{table}[H]
\centering
\caption{Performance on \textit{InternVL-v1.5-7B}.}
\label{tab: InternVL}
\vspace{-0.1in}
\scalebox{0.9}{
\begin{tabularx}{\columnwidth}{Xcccc}
\toprule
Method & T-2  & S-4  & NH  & IR  \\ \midrule
\rowcolor{gray!16}
\textbf{Full Cache}  & 19.2 & 19.1 & 11.1 & 0.0 \\
\midrule
$\textbf{H}_{\textbf{2}}\textbf{O}$        & 20.0 & 19.6 & 3.9  & 0.5 \\
\textbf{SnapKV}      & 19.9 & 19.4 & 4.1  & 0.2 \\
\textbf{RoCo}  & 20.0 & 19.6 & 3.9  & 0.5 \\
\midrule
\rowcolor{green!16}
\textbf{LOOK-M} & \textbf{22.0} & \textbf{22.9} & \textbf{10.9} & \textbf{0.5} \\ \bottomrule
\end{tabularx}
}
\end{table}

\begin{table}[H]
\centering
\caption{Performance on \textit{MobileVLM\_V2-3B}.}
\label{tab: MobileVLM}
\vspace{-0.1in}
\scalebox{0.9}{
\begin{tabularx}{\columnwidth}{Xcccc} 
\toprule
Method & T-2  & S-4  & NH  & IR  \\ \midrule
\rowcolor{gray!16}
\textbf{Full Cache}  & 46.2 & 33.0 & 10.6 & 4.7 \\
\midrule
$\textbf{H}_\textbf{2}\textbf{O}$        & 46.4 & 28.2 & 4.2  & 4.5 \\
\textbf{SnapKV}      & 46.4 & 27.2 & 4.4  & 4.7 \\
\textbf{RoCo}   & 46.6 & 28.9 & 4.2  & 4.7 \\
\midrule
\rowcolor{yellow!16}
\textbf{LOOK-M} & \textbf{47.0} & \textbf{32.8} & \textbf{10.3} & \textbf{4.8} \\ \bottomrule
\end{tabularx}}
\end{table}

\subsection{Performance on Different  Architectures}
To validate the effectiveness of the LOOK-M method across various architectures, we tested its performance not only on the LLaVA architecture but also on mobileVLM and InternVL. We selected several representative multimodal long-context subtasks from MileBench, including T2 (Temporal Multi-image), S-4 (Semantic Multi-image), NH (Needle in a Haystack), and I (Image Retrieval). From the results presented in Tables~\ref{tab: InternVL} and ~\ref{tab: MobileVLM}, LOOK-M consistently outperformed traditional eviction-based methods, including $\text{H}_{\text{2}}\text{O}$, SnapKV, and RoCo. Notably, in both architectures, LOOK-M demonstrated significant advantages over other baselines in Needle in a Haystack, the multimodal long-context retrieval task. This confirms that LOOK-M’s pivotal merging strategy effectively preserves key multimodal representations while compressing the KV cache for accurate information retrieval, with minimal information loss compared to Full Cache.

\subsection{Influence of Various Cache Budgets }
\begin{figure*}[h]
\centering
\includegraphics[width=\linewidth]{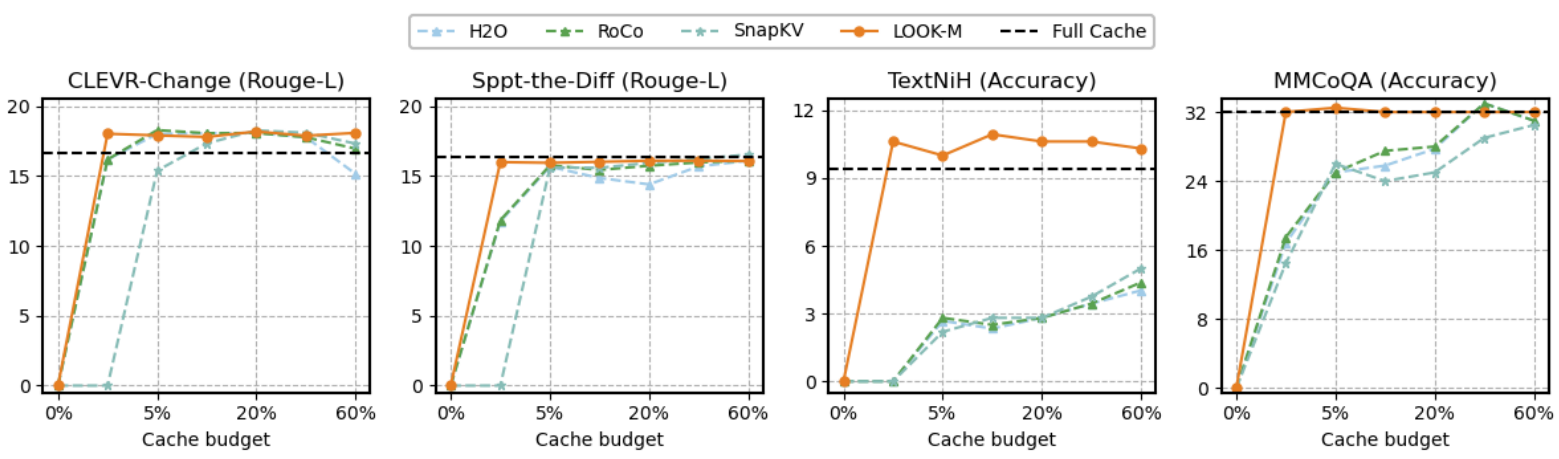}
\caption{Influence of Various Cache Budgets on Performance.}
\label{fig:cachebudget}
\end{figure*}

In this section, we assess the efficiency of the LOOK-M strategy under varying KV cache budgets by conducting standardized tests on the LLaVA-v1.5-7B model and four subtasks: CLEVR-Change, Spot-the-Diff, TextNiH, and MMCoQA. As depicted in Figure~\ref{fig:cachebudget}, LOOK-M approaches Full Cache performance even with an extreme KV cache compression of 5\%, especially using the text-prior pivotal merging strategy. Particularly in the TextNiH and MMCoQA tasks, it consistently outperforms the baselines regardless of compression rate. These results demonstrate that, despite the redundancy of tokens within the multimodal long-context KV cache, traditional algorithms' maximal compression often results in considerable loss of information. Conversely, LOOK-M effectively preserves critical information with a minimal KV budget, with its merging strategy significantly reducing context loss.

\begin{figure*}[ht]
\vspace{-0.2mm}
\centering
\includegraphics[width=\linewidth]{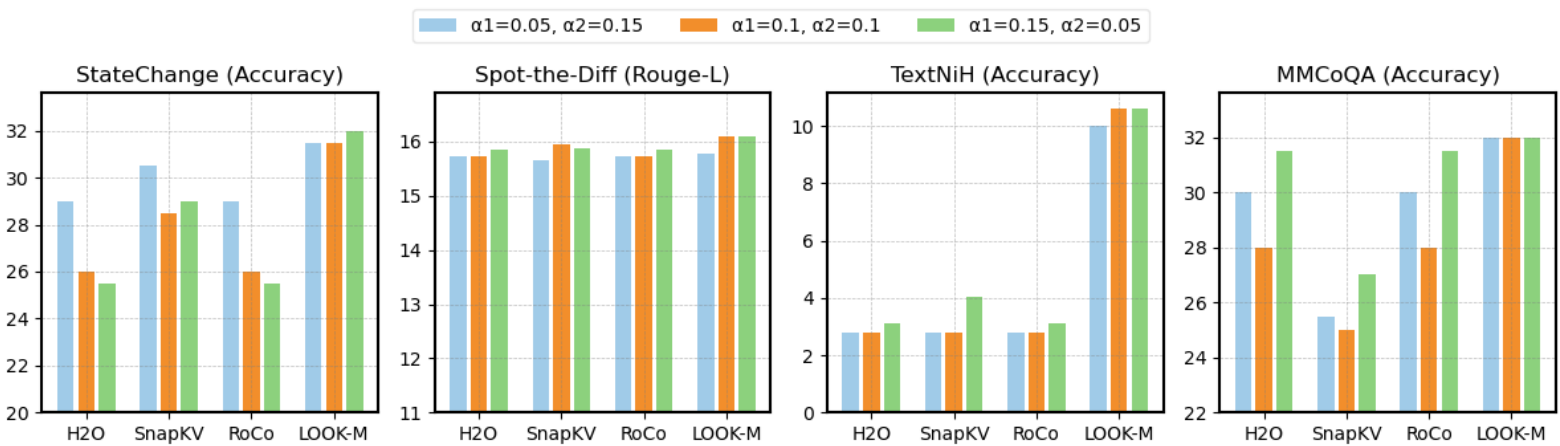}
\caption{Impact of Different Compression Ratio Proportion.}
\label{fig:cacheratio}
\vspace{-0.2mm}
\end{figure*}

\subsection{Hyperparameter Analysis on $\alpha_{1}$ and $\alpha_{2}$}

To evaluate the impact of varying the recent token ratio ($\alpha_1$) and important token ratio ($\alpha_2$) on model performance, we conducted tests across four different datasets using the LLaVA-v1.5-7B model. As shown in Figure~\ref{fig:cacheratio}, LOOK-M consistently outperformed other baselines under different settings of $\alpha_1: \alpha_2$ ratios, particularly showing significant advantages in the StateChange and MMCoQA datasets at every ratio.  Furthermore, we observed that for LOOK-M, a higher important token ratio $\alpha_2$ correlates with improved performance, suggesting that when less context information is discarded, the merging strategy is more effective.

\subsection{Efficiency Analysis}
In this section, we analyze the efficiency of our proposed LOOK-M method, as illustrated in Table \ref{tab:model_performance}. We compare the decoding speed and memory usage of model inference with and without our LOOK-M method. To ensure the robustness of our results, the tests for decoding latency and GPU memory usage were specifically conducted on 20 randomly selected data entries from the MileBench dataset. Additionally, the speed tests were performed using RTX 3090 $\times$ 1. 

\begin{table}[ht]
\centering
\captionsetup{justification=centering}
\caption{Model Speed and KV Cache GPU Memory Usage.}
\scalebox{0.8}{
\begin{tabular}{lccc}
\toprule
Method & Budge & Decoding Latency & GPU Memory \\ 
\midrule
Full Cache & 100\% &  28.16 ms/token & 1.52 GiB \\ 
LOOK-M  & 20\% &  20.98 ms/token & 0.32 GiB \\
LOOK-M  & 5\% & \textbf{18.22 ms/token} & \textbf{0.13 GiB} \\
\bottomrule
\end{tabular}
}
\label{tab:model_performance}
\end{table}
\vspace{-0.1in}
As we can observe from Table \ref{tab:model_performance}, the decoding latency of our compressed model is significantly lower than that of the model retaining the full cache, with the advantage becoming more pronounced in the generation of long texts. This highlights the efficiency of our method in tasks involving long text generation. Additionally, we analyzed the speed and GPU memory usage of the KV Cache under two budget scenarios: 20\% and 5\%, based on the mean values from the inference process of 20 randomly sampled data points (as illustrated in Table \ref{tab:model_performance}, Our findings indicate that the average GPU memory consumption is nearly proportional to the cache budget. At a 20\% KV Cache budget, memory usage during the decode stage is reduced by approximately 80\% compared to a Full Cache scenario. Furthermore, an increase in the compression ratio significantly reduces decoding latency, thus enhancing the decode stage's efficiency and demonstrating the effectiveness of our compression method.

\section{Conclusion}

In this work, we propose \textbf{L}ook-\textbf{O}nce \textbf{O}ptimization in \textbf{K}V for Efficient \textbf{M}ultimodal long-context inference (\textbf{LOOK-M}),
the first framework is specifically designed to manage multimodal KV caches in multimodal large language models (MLLMs) efficiently. LOOK-M integrates a novel KV cache eviction strategy with innovative merging techniques, such as averaged, weighted, and pivotal merging, to maintain essential contextual information without the need for fine-tuning. Our findings reveal that the framework not only preserves the quality of generation in multimodal long-text scenarios but also ensures robust performance under significant KV cache compression. Observations indicate that LOOK-M prioritizes text over visual inputs during prompt prefilling, leading to the development of a text-prior method that further optimizes KV cache compression. Looking ahead, we plan to expand LOOK-M's capabilities by incorporating additional compression techniques like quantization, distillation, and efficient attention mechanisms to enhance both efficiency and efficacy.

\section{Limitation}

The constraints of our work lie in the fact that we have used plain multimodal large language models (MLLMs) without incorporating advanced compression techniques such as quantization, distillation, and efficient attention mechanisms. In our future research, we plan to explore methods to achieve the most extreme level of KV cache compression. 
Additionally, by optimizing the multimodal KV cache, our technique allows MLLMs to run on resource-limited devices like smartphones and laptops while maintaining inference accuracy. This capability supports diverse applications, including healthcare~\cite{wan2024med, wan2024electrocardiogram, wan2022g, liu2024etp, liu2024benchmarking, liu2024zero, zheng2024structure}, math~\cite{DBLP:journals/corr/abs-2110-14168, xiong2022self}, optimization~\cite{liang2020large, liang2020many}, and recommendation~\cite{wan2023spatio}, and aids in developing MLLMs for various technological environments. However, improper application of this compression method, particularly at high compression ratios, may reduce performance and affect functionality.

\bibliography{custom}

\newpage
\appendix

\section{Appendix}
\label{sec:appendix}
\subsection{Details of MileBench}
MileBench~\cite{Song2024MileBenchBM} dataset is the first benchmark specifically designed to test the Multimodal Long-context capabilities of MLLMs. Milebench primarily includes 6,440 multimodal long-text samples, which are composed of 21 existing or self-constructed datasets, with an average of 15.2 images and 422.3 words per sample. It composed of two primary subsets: \textbf{\textit{Realistic Evaluation}} and \textbf{\textit{Diagnostic Evaluation}}. 

\noindent\textbf{\textit{Realistic Evaluation}} component challenges MLLMs to manage tasks within multimodal long-context situations, underscoring the models' ability to understand and reason through prolonged multimodal contexts. 

\noindent\textbf{\textit{Diagnostic Evaluation}} requires MLLMs to extract information from the given context, accentuating the models' skills in long-distance information retrieval and the removal of distractors. 

The comprehensive classification of Milebench is presented in Table \ref{tab:taxonomy_details}.

\subsection{Performance under extreme compression ratio}
We evaluate the performance of various KV Cache compression strategies at compression ratios exceeding 80\%, as detailed in the main text. Notably, Table \ref{tab:high ratio} reveals that at an extreme compression ratio of 99\%, our method, \textbf{LOOK-M}, exhibits a significant advantage over competing methods. It consistently maintains performance across the vast majority of sub-tasks, closely matching the results achieved using a Full Cache. This outcome not only underscores the robustness of our method at high compression ratios but also its superior ability to sustain performance relative to other approaches.

\begin{table*}[t!]
\centering
\caption{Detailed Taxonomy of MileBench.~\cite{Song2024MileBenchBM}}
\resizebox{\textwidth}{!}{%
    \begin{tabular}{l|l|c|c}
        \toprule
        \textbf{Category} & \textbf{Task} & \textbf{Dataset} & \textbf{Metric} \\
        \midrule
        \rowcolor{gray!17} \multicolumn{4}{c}{\textbf{\textit{Realistic Evaluation}}} \\
        \midrule
         & \textbf{Action Understanding and} & Action Localization & Accuracy \\
         & \textbf{Prediction (T-1)} & Action Prediction & Accuracy \\
         & & Action Sequence & Accuracy \\
        
        \cmidrule(lr){2-2} \cmidrule(lr){3-3} \cmidrule(lr){4-4}
         & \textbf{Object and Scene} & Object Existence & Accuracy \\
         & \textbf{Understanding (T-2)} & Object Interaction & Accuracy \\
        \textbf{Temporal} & & Moving Attribute & Accuracy \\
        \textbf{Multi-image} & & Object Shuffle & Accuracy \\
         
        \cmidrule(lr){2-2} \cmidrule(lr){3-3} \cmidrule(lr){4-4}
         & \textbf{Visual Navigation and} & Egocentric Navigation & Accuracy \\
         & \textbf{Spatial Localization (T-3)} & Moving Direction & Accuracy \\
         
        \cmidrule(lr){2-2} \cmidrule(lr){3-3} \cmidrule(lr){4-4}
         & \textbf{Counterfactual Reasoning} & Counterfactual Inference & Accuracy \\
         & \textbf{and State Change (T-4)} & State Change & Accuracy \\
         & & Character Order & Accuracy \\
         & & Scene Transition & Accuracy \\
         
        \midrule
         & \textbf{Knowledge Grounded QA (S-1)} & Webpage QA & Accuracy \\
         & & Textbook QA & Accuracy \\
         & & Complex Multimodal QA & Accuracy \\
         & & Long Text with Images QA & Accuracy \\
         
        \cmidrule(lr){2-2} \cmidrule(lr){3-3} \cmidrule(lr){4-4}
         & \textbf{Text-Rich Images QA (S-2)} & Slide QA & Accuracy \\
        \textbf{Semantic} & & OCR QA & Accuracy \\
        \textbf{Multi-image} & & Document QA & Accuracy \\
         
        \cmidrule(lr){2-2} \cmidrule(lr){3-3} \cmidrule(lr){4-4}
         & \textbf{Visual Relation Inference (S-3)} & Visual Change Captioning & ROUGE-L \\
         & & Visual Relationship Expressing & ROUGE-L \\
            
        \cmidrule(lr){2-2} \cmidrule(lr){3-3} \cmidrule(lr){4-4}
         & \textbf{Dialogue (S-4)} & Multimodal Dialogue & Accuracy \\
         & & Conversational Embodied Dialogue & ROUGE-L \\
        \cmidrule(lr){2-2} \cmidrule(lr){3-3} \cmidrule(lr){4-4}
         & \textbf{Space Understanding (S-5)} & Space Understanding & Accuracy \\
        \midrule
        \rowcolor{gray!17} \multicolumn{4}{c}{\textbf{\textit{Diagnostic Evaluation}}}\\
        
        \midrule
        \textbf{Needle In} & \textbf{Text Needle (N-1)} & Text Needle In A Haystack & Accuracy \\
        
        \cmidrule(lr){2-2} \cmidrule(lr){3-3} \cmidrule(lr){4-4}
        \textbf{A Haystack} & \textbf{Image Needle (N-2)} & Image Needle In A Haystack & Accuracy \\
        \midrule
        \textbf{Image Retrieval} & \textbf{Image Retrieval (I-1)} & Image Retrieval & Accuracy \\
        \bottomrule
    \end{tabular}
}
\label{tab:taxonomy_details}
\end{table*}

\begin{table*}[t]
\centering
\caption{Comparative Performance of Different Strategies at Maximum Compression Rate(99\%) on LLaVA-v1.5-7B}
\label{tab:high ratio}
\small 
\begin{tabularx}{\textwidth}{l|*{11}{>{\centering\arraybackslash}X}}
\toprule
Method & T-1 & T-2 & T-3 & T-4 & S-1 & S-2 & S-3 & S-4 & S-5 & NH & IR \\ 
\midrule
\rowcolor{gray!16}
\textbf{Full Cache} & 40.0 & 46.0 & 32.2 & 37.8 & 56.9 & 33.3 & 12.6 & 23.4 & 60.5 & 4.7 & 4.3 \\
\midrule
$\textbf{H}_\textbf{2}\textbf{O}$ & 36.5 & 46.0 & 25.0 & 31.5 & 36.4 & 23.0 & 9.4 & 9.4 & 51.5 & 0.0 & 3.3 \\ 
\textbf{SnapKV} & 38.8 & 45.1 & 26.5 & 34.1 & 38.4 & 26.0 & 0.0 & 9.6 & 58.0 & 0.0 & 3.5 \\ 
\textbf{RoCo} & 36.5 & 46.1 & 25.2 & 32.5 & 36.4 & 23.0 & 9.4 & 9.2 & 52.5 & 0.0 & 3.1 \\ 
\midrule
\rowcolor{green!20}
\textbf{LOOK-M} & \textbf{40.3} & \textbf{46.1} & \textbf{32.5} & \textbf{40.0} & \textbf{57.0} & \textbf{33.7} & \textbf{12.8} & \textbf{24.0} & \textbf{60.0} & \textbf{5.3} & \textbf{3.7} \\ 
\bottomrule
\end{tabularx}
\end{table*}

\end{document}